\documentclass{article} 
\usepackage[final]{colm2026_conference}

\usepackage{microtype}
\usepackage{hyperref}
\usepackage{url}
\usepackage{booktabs}
\usepackage{amsmath}
\usepackage{algorithm}
\usepackage{amssymb}
\usepackage{amsthm}
\usepackage{mathtools}
\usepackage{bm}

\usepackage{listings}
\usepackage{xcolor}
\usepackage{cleveref}

\usepackage{algorithm}
\usepackage{algpseudocode}
\algrenewcommand\algorithmicrequire{\textbf{Input:}}
\algrenewcommand\algorithmicensure{\textbf{Output:}}
\theoremstyle{definition} 
\DeclareMathOperator*{\argmin}{argmin}

\usepackage{lineno}
\usepackage{multirow}

\usepackage{authblk}
\definecolor{darkblue}{rgb}{0, 0, 0.5}
\hypersetup{colorlinks=true, citecolor=darkblue, linkcolor=darkblue, urlcolor=darkblue}
\usepackage{graphicx}
\usepackage{wrapfig}
\usepackage[table]{xcolor}

\title{When Prompts Interact: Assessing Prompt Arithmetic for Deconfounding under Distribution Shift}

\author[1]{Zhecheng~Sheng}
\author[2]{Yongsen~Tan}
\author[3]{Xiruo~Ding}
\author[2]{Trevor~Cohen}
\author[1]{Serguei~Pakhomov}

\affil[1]{University of Minnesota}
\affil[2]{University of Washington}
\affil[3]{Stanford University}

\begin{document}

\ifcolmsubmission
\linenumbers
\fi

\maketitle

\begin{abstract}
In classification tasks, models may rely on confounding variables to achieve strong in-distribution performance, capturing spurious features that fail under distribution shift. This shortcut behavior leads to substantial degradation in out-of-distribution settings. Task arithmetic offers a potential solution by removing unwanted signals via subtraction of secondary model updates, but it typically requires full fine-tuning, which is computationally expensive. Prompt tuning provides a parameter-efficient alternative by adapting models through a small set of trainable virtual tokens. Task arithmetic on the resulting prompts presents an appealing alternative to operations on entire models, but the extent to which this approach can limit reliance on spurious features remains to be established. In this work, we study whether composing soft prompts through task arithmetic improves robustness to confounding shifts. We propose Hybrid Prompt Arithmetic (HyPA), which combines task prompts with linearized confounder prompts to counteract spurious correlations. Across multiple benchmarks, HyPA consistently improves the robustness-performance trade-off relative to prompt-arithmetic baselines under distribution shift. We further analyze how HyPA affects hidden representations and find evidence consistent with it mitigating confounding either by reducing the influence of confounder signals on predictions or by suppressing them in the representation. These results establish HyPA as a parameter-efficient and promising approach for improving robustness under confounding shifts in the evaluated setting. Our codebase is available at \url{https://github.com/zcsheng95/prompt_arithmetics}.
\end{abstract}

\section{Introduction}
Machine learning models trained via empirical risk minimization (ERM) have been shown to exploit confounding features that are predictive in the training distribution but not causally relevant to the target task. This phenomenon is known as shortcut learning \citep{geirhos2020shortcut}. While such shortcuts can yield strong in-distribution performance, reliance on spurious correlations leads to substantial degradation under distribution shift \citep{recht2019imagenet, koh2021wilds}.
Prior mitigation strategies either introduce training-time constraints in the loss function which encourage a model to learn invariant features across confounding groups or environments \citep{sagawa2019distributionally, arjovsky2019invariant}, or apply post-hoc corrections to a biased predictor \citep{dingTailoringTaskArithmetic2025, sheng-etal-2025-mitigating}. Our work follows the latter line and develops a prompt-based debiasing method for transformers \citep{vaswani2017attention} without requiring modifications to the original training pipeline.

Specifically, we combine prompt tuning and task arithmetic to mitigate confounding effects in model predictions. While task arithmetic enables composition of model updates to modify model capabilities \citep{Ilharco2022EditingMW}, it exhibits a trade-off between expressivity and disentanglement. In the context of a classification task with a known confounding variable, standard non-linear fine-tuning achieves strong performance on the primary objective but entangles spurious confounder-related features, whereas linearized fine-tuning \citep{ortiz2023task} improves disentanglement at the cost of reduced expressivity.

To address this limitation, we propose \textbf{Hybrid Prompt Arithmetic (HyPA)}, a two-phase framework that combines task vectors obtained from both non-linear and linearized prompt tuning. The key idea is to anchor the model in an optimized task solution and then estimate confounder-specific directions within a locally linearized parameter space. By composing these components through prompt arithmetic, HyPA effectively suppresses spurious correlations while preserving task-relevant representations. Empirically, we show that HyPA improves out-of-distribution robustness across multiple benchmarks while maintaining in-distribution performance.

\section{Problem Setup and Preliminaries}
We begin by formalizing the distribution shift problem setup and introducing the key components of our method. Consider a dataset $\mathcal{D}$ with $(X, Y, Z) \sim \mathcal{D}$, where $X \in \mathcal{X}$ denotes the input text, $Y \in \mathcal{Y}$ the task label, and $z \in \mathcal{Z}$ the confounding attribute. We assume that $Z$ participates in the underlying data-generating process of both $X$ and $Y$, corresponding to the causal structure \citep{pearlCausality2009}: $X \leftarrow Z \rightarrow Y$. Confounding shift occurs when $P(Y|Z)$ at test time differs from its training-time value. Given access to $(X, Y, Z)$ during training, our goal is to learn a discriminative function $f_\theta : \mathcal{X} \rightarrow \mathcal{Y}$ that is robust to confounding shift at test time, where $Z$ may or may not be observed. Intuitively, this amounts to regularizing the learned model to avoid relying on spurious pathways from $Z$ to $Y$ when making predictions.
\paragraph{Confounding Shift.} Confounding shift \citep{landeiro2018robust} refers to a setting in which the conditional distribution of the label given the confounder differs between training and test environments, i.e., $P_{\text{train}}(Y \mid Z) \neq P_{\text{test}}(Y \mid Z)$, thereby violating the \textit{i.i.d.} assumption commonly adopted in supervised learning. When confounding shift is present, a predictor $f_\theta$ that exploits correlations between $Y$ and $Z$ to learn the association between $Y$ and $X$ (via features in $X$ that are also associated with $Z$) during training may experience performance degradation at inference time (i.e., during testing on data in which $Z$ may be unobserved or distributed substantially differently from $Z$ in the training data.). To quantify the degree of confounding shift, \citet{ding2024backdoor} introduce an auxiliary variable $\alpha$ defined as
\begin{equation}
\alpha = \frac{P(Y = 1 \mid Z = 1)}{P(Y = 1 \mid Z = 0)}.
\end{equation}
When $\alpha > 1$, positive labels are more prevalent in the subgroup $Z = 1$, and conversely when $\alpha < 1$. In other words, $\alpha > 1$ indicates that the confounder and the positive class label are found together more frequently than the positive class label without the confounder, and vice versa for $\alpha < 1$. Confounding shift occurs when the value of $\alpha$ differs between training and test time. Note this formulation is restricted to binary outcomes $Y$ and binary confounders $Z$.

\paragraph{Prompt Tuning.} Prompt tuning \citep{lester-etal-2021-power} is a parameter-efficient adaptation method that learns a set of additional parameters to condition a frozen base model on specific downstream tasks. Let $f_{\theta}$ denote a pretrained model with fixed parameters $\theta$, $x = (x_1, x_2, \ldots, x_n)$ be an input token sequence, and $y$ be the corresponding task label. Prompt tuning introduces a sequence of trainable soft prompt embeddings $\mathbf{P} = (p_1, p_2, \ldots, p_m)$, where $p_i \in \mathbb{R}^d$ and $d$ is the dimensionality of the model’s hidden state. With the soft prompt embeddings prepended to the input embeddings, the optimization objective becomes
\begin{equation}
\mathbf{P^*} = \argmin\limits_{\mathbf{P}}\mathbb{E}_{(x,y)}[\mathcal{L}(f_{\theta}(x; \mathbf{P}), y)],
\end{equation}
where $\theta$ is fixed and only $\mathbf{P}$ is optimized via backpropagation. $\mathcal{L}$ is an arbitrary loss function.

\paragraph{Task Vectors \& Task Arithmetic.} In contrast to prompt tuning in which the model's parameters $\theta$ are frozen, the task vector approach consists of deriving a task vector $\tau_t$ that represents the element-wise difference between the fine-tuned model weights $\theta_t$ for task $t$ and the pretrained weights $\theta_0$, where both $\theta_0 \in \mathbb{R}^d$ and $\theta_t \in \mathbb{R}^d$. \citet{Ilharco2022EditingMW} proposed methods to post-edit the pretrained model weights through task arithmetic, which linearly combines a set of task vectors (in this case, the ``vectors'' are the model weights in totality) from different tasks and modifies the performance of each task independently. Formally, task arithmetic can be expressed using the following formula:
\begin{equation}
   \tau_t = \theta_t - \theta_0, \quad \theta_{\text{new}} = \theta_0 + \sum_{t=1}^{T}\lambda_t\tau_t,
\end{equation}
where $\lambda_t$ is a scalar coefficient for task $t$, which can be negative when the goal is to remove an undesirable effect. \citet{ortiz2023task} also note that, for a task vector set $\mathrm{T} = \{\tau_t\}_{t \in [T]}$, their task supports $\mathcal{D} = \{\mathcal{D}_{t} \subset \mathcal{X}\}$ should not overlap between different tasks $t$. However, in our experimental setup, we use the same input $x$ to compute different task vectors based on their different labels.

\paragraph{Linearization.}
Prior work on the Neural Tangent Kernel (NTK) \citep{jacot2018neural} shows that a wide neural network can be locally approximated by a linear function around its initialization using a first-order Taylor expansion
\begin{equation}\label{eq:linear}
f_{\theta}(x) \approx f_{\theta_0}(x)
+ \nabla_{\theta} f_{\theta_0}(x)^\top (\theta - \theta_0) =  f_{\theta_0}(x)
+ \nabla_{\theta} f_{\theta_0}(x)^\top \tau.
\end{equation}
Under this approximation, the parameter-to-function mapping is linear. This regime holds for sufficiently wide networks with small parameter updates, which is common in modern overparameterized models. The task vector $\tau$ in Equation \ref{eq:linear} can be optimized directly. 

Although NTK provides a tractable linearization of neural network training dynamics, it may inadequately model feature learning, which can result in reduced task performance \citep{chizat2019lazy, seleznova2022neural}.

\section{Prompt Arithmetic in Causal Language Models}

Our objective is to learn a classifier $f$ satisfying $f(X) \perp Z \mid Y$, such that predictions remain invariant to shifts in $P(Y|Z)$ at test time. This requires the model to base its predictions on features $X^{\perp}_{Z}$ that are independent of $Z$. To this end, we adopt a post-hoc strategy inspired by task arithmetic, which posits that a model’s behavior across different tasks can be approximated as additive under certain conditions \citep{ortiz2023task}.
Motivated by the effectiveness of task arithmetic, we propose methods that selectively eliminate confounder-related signals through linear combinations of task vectors that represent the confounder and the target tasks. Most prior work on task arithmetic focuses on manipulating task vector $\tau$ using the full set of model parameters. In contrast, we propose performing task arithmetic only within the soft prompt space, a lightweight module inserted at the input embedding layer. This method formulation improves scalability and obviates the need for additional task-specific submodules such as classification heads (see Appendix~\ref{sec:frozen_lm} for details).

\subsection{Hybrid Prompt Arithmetic (HyPA)}

\paragraph{Prompt Arithmetic (PA).}
We restrict the tunable parameters in task arithmetic to the soft prompt embeddings, denoted by $\theta \coloneq \mathbf{P}$. Prompt Arithmetic is then defined as
\begin{equation}
\begin{aligned}
\tau_t = \mathbf{P}_t - \mathbf{P}_0, \quad
\mathbf{P}_{\text{new}} = \mathbf{P}_0 + \sum_{t=1}^{T} \lambda_t \tau_t,
\end{aligned}
\end{equation}
where $\mathbf{P}_0$ denotes the initialized soft prompt embeddings, in contrast to the pretrained model weights $\theta_0$ used in standard task arithmetic. This set up is conceptually the same as the task prompt vector concept proposed in \cite{belanec2025taskpromptvectorseffective}.

Prior work shows that task vectors can be obtained via either full non-linear fine-tuning \citep{Ilharco2022EditingMW} or a linearized approximation \citep{ortiz2023task}, referred to as linear fine-tuning. Linear fine-tuning promotes better weight disentanglement of parametric function $f$, which facilitates effective task arithmetic, but it reduces expressivity because restricting updates to a linear regime can degrade single task performance compared to non-linear fine-tuning \citep{jin2025finetuningattentionmodulesonly}.

\paragraph{Hybrid Prompt Arithmetic (HyPA).}
\begin{wrapfigure}{r}{0.40\textwidth}
    \centering
    \includegraphics[width=\linewidth]{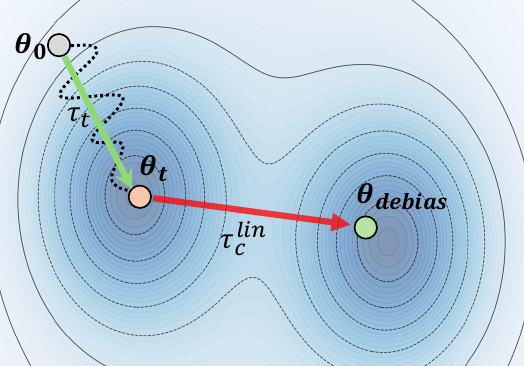}
    \vspace{-3mm}
    \caption{Traversing the task loss landscape along a linearized direction to achieve robustness while maintaining performance.}
    \vspace{-3mm}
    \label{fig:illustration}
\end{wrapfigure}

To balance these trade-offs, we propose \textbf{Hybrid Prompt Arithmetic (HyPA)}, 
a two-phase training strategy that combines task vectors from the original non-linear model and its linearized approximation to derive a de-confounded model. Figure~\ref{fig:illustration} illustrates the overall procedure for obtaining a debiased model from initialized or pretrained weights.  The scaling factor $\lambda$ controls the strength of confounder removal in the final model $f_{\theta_{\text{debias}}}$.

Recall that fine-tuning is performed using soft prompts, so the effective tunable parameters are $\theta \coloneq \mathbf{P}$. In the first phase, we fine-tune the model starting from pretrained weights $\mathbf{P}_0$ using the full non-linear network to obtain a task-specific model $\mathbf{P}_t$ and its corresponding task vector $\tau_t = \mathbf{P}_t - \mathbf{P}_0$. In the second phase, we constrain confounder fine-tuning to the \textit{tangent space} around the task-specific weights $\mathbf{P}_t$, yielding a confounder task vector obtained from a linearized model:
\begin{equation}
\begin{aligned}
f_{\theta}(x;\mathbf{P}_t) = f_{\theta}(x;\mathbf{P}_0 + \tau_t), \quad
f_{\theta}^{\mathrm{lin}}(x;\mathbf{P}_t + \tau_c^{\mathrm{lin}})
&=
f_{\theta}^{\mathrm{lin}}(x;\mathbf{P}_t)
+ \nabla_{\mathbf{P}} f_{\theta}(x;\mathbf{P}_t)^{\top} \tau_c^{\mathrm{lin}} .
\end{aligned}
\end{equation}
Here, $\tau_t$ denotes the task vector obtained from non-linear fine-tuning, while $\tau_c^{\mathrm{lin}}$ denotes the confounder task vector derived from the linearized model $f^{\mathrm{lin}}$.

Task arithmetic is then applied to negate confounder signals by combining the confounder task vector with the anchored task-specific weights. This operation aims to remove spurious correlations while preserving task-relevant features learned during non-linear fine-tuning:
\begin{equation}\label{eq:p_debias}
\mathbf{P}_{\text{debias}} = \mathbf{P}_0 + \tau_t + \lambda \cdot \tau_c^{\mathrm{lin}} .
\end{equation}
The hyperparameter $\lambda$ determines the extent to which $\tau_c^{\mathrm{lin}}$ influences the behavior of the debiased model.  The de-confounded model output is obtained by $f_{\theta}(x; \mathbf{P}_{\text{debias}})$ with $\mathbf{P}_{\text{debias}}$ being the plug-in adapter for the base model. The algorithm is formalized in Appendix~\ref{alg:hypa}.

\section{Experiments}
\paragraph{Dataset.} We manipulate the strength of spurious correlation, characterized by $P(Y \mid Z)$, through controlled sampling at both training and test time to construct experimental datasets. Specifically, we fix the marginal distributions $P_Y$ and $P_Z$ to be uniform for both training and testing, thereby isolating the effect of confounding shift. We vary the parameter $\alpha$ during training and testing to induce different levels of confounding shift. In our experiments, we set $\alpha_{\mathrm{train}} \in \{0.2, 5.0\}$ to construct training datasets with strong spurious correlations. At test time, we generate a sequence of $\alpha$ values that are uniformly spaced in the $\log$ scale (since $\alpha$ is defined as a ratio) and evaluate model performance across these shifts.

Through the DeconDTN Toolkit \citep{pmlr-v333-tan26a}, we construct data from five data sources for text categorization, ranging from clinical text to general online comments, to examine the generalizability of the proposed method. All the datasets focus on a binary prediction task from text input with a binary confounder label. Detailed descriptions and statistics of these datasets can be found in Appendix~\ref{sec:dataset}.

\paragraph{Baselines.} 
In our main results, we compare \textbf{HyPA} with two baseline variants: (i) \textbf{PA}, which derives both task and confounder vectors via non-linear fine-tuning, and (ii) \textbf{LinPA}, which derives both vectors from linearized models. For the above two baselines, we adopt the original settings in task arithmetic where both task and confounder models are fine-tuned from the same initialization point. We use GPT-2 \citep{radford2019language} as the base model and a prompt length of $100$ in the main results. 
We demonstrate the comparison results with a broader selection of baseline methods in Section~\ref{sec:other_baseline}.

\paragraph{Model Selection.} 
Since test-time shift is unknown in practice, we tune the hyperparameter $\lambda$ using the worst-group AUPRC on a held-out validation set, which has the same distribution as the training set. When $\alpha_{\mathrm{train}} < 1$, the worst-performing group corresponds to samples with $Z=1$, and when $\alpha_{\mathrm{train}} > 1$, the worst-performing group corresponds to samples with $Z=0$. This aligns with previous practice when confounder attributes are available during training and validation \citep{sagawa2019distributionally, gulrajani2020search,idrissi2022simple, yang2023change}.

\paragraph{Evaluation Metrics.} We use four metrics to evaluate the performance of the debiased model under varying degrees of confounding shift. The Out-of-Distribution (OOD) score measures performance under extreme confounding shifts at test time. The Absolute Slope quantifies performance variation across different levels of confounding shift, while the Adjusted Integral jointly captures both performance and robustness in a single measure.

\textbf{OOD Score.} We evaluate the model $performance$ on the classification task of interest when the confounder distribution in the test set is exactly the reciprocal of the training distribution, representing a case of severe confounding shift:
\begin{equation}
    \text{OOD Score} = \mathbf{s}_{\alpha}, \quad \text{where } \alpha=\frac{1}{\alpha_\text{train}} 
\end{equation}
\textbf{Absolute Slope.} Given a collection of data points generated from a list of $\alpha_{\text{test}}$, the Absolute Slope measures the $robustness$ of the model's performance on this test set collection. It is quantified as the absolute value of the coefficient from the linear regression fitted between the model performance score $\mathbf{s}$ and $\log(\alpha_{\text{test}})$:
\begin{equation}
    \mathbf{s} = \beta_0 + \beta \times \log(\alpha_{\text{test}}),
\end{equation}
where $\beta_0$ is the intercept and $|\beta|$ is the measure of interest for robustness. Intuitively, a flatter regression line signifies greater robustness of the model under confounding shift.

The two metrics above each assess a single aspect of model behavior separately: $performance$ and $robustness$, respectively. To characterize the interaction between both of these aspects of model behavior,  we introduce two metrics that jointly capture both performance and robustness: Area Under the Shift Curve (AUSC) and Adjusted Integral (AI).

\textbf{Area Under Shift Curve (AUSC).} The Area Under the Shift Curve  captures the average of the AUPRCs across different test-time confounding distribution shifts (i.e., mean AUPRC over the support of $alpha$). Larger AUSC values indicate consistently higher AUPRCs under various distribution shifts.

\begin{equation}
    \text{AUSC}
     = \frac{1}{\log(a) - \log(b)}
    \int_{\log(b)}^{\log(a)}
    \mathbf{s}_\alpha
    \, d(\log \alpha)
\end{equation}

\textbf{Adjusted Integral.} This metric is defined as the absolute difference between out-of-distribution (shift) performance and in-distribution performance, augmented by a penalty term that compensates for deficiencies in in-distribution performance.

\begin{equation}
\text{Adjusted Integral}
=
\frac{1}{\log(a) - \log(b)}
\int_{\log(b)}^{\log(a)}
\left|
\mathbf{s}_{\text{ID}} - \mathbf{s}_\alpha
\right|
\, d(\log \alpha)
\;+\;
\left(1 - \mathbf{s}_{\text{ID}}\right),
\end{equation}
where $\mathbf{s}_{\text{ID}}$ denotes the in-domain model performance (i.e., at $\alpha = \alpha_{\text{train}}$), and $a$ and $b$ denote the maximum and minimum values in the test-time shift set ${\alpha_{\text{test}}}$, respectively. The term $1 - \mathbf{s}_{\text{ID}}$ penalizes poor in-domain performance.


For all metrics above, we use Area Under the Precision and Recall Curve (AUPRC) as the performance measure $\mathbf{s}$ and evaluate shifts using uniform spacing in $\log_{10}\alpha$.
\section{Main Results}
Table~\ref{tab:main_results} shows the performance of PA, LinPA, and HyPA across five benchmarks under two training shift settings ($\alpha_{\text{train}} = 0.2$ and $\alpha_{\text{train}} = 5.0$). Overall, HyPA consistently achieves the strongest joint performance, obtaining the best or near-best results in terms of AUSC and Adjusted Integral across datasets and shift settings.
\vspace{-3mm}
\begin{table}[ht]
\centering
\caption{Performance–robustness trade-offs across five datasets under two strengths of training-time spurious correlation. Values are means with standard deviations in parentheses over five random seeds; the best result in each setting is bolded.}
\resizebox{\textwidth}{!}{%
\begin{tabular}{lc*{5}{>{\centering\arraybackslash}p{2.8cm}}} 
\toprule
& & \textbf{SHAC} & \textbf{MIMIC} & \textbf{Hate Speech} & \textbf{Civil Comments} & \textbf{Amazon Reviews} \\
\cmidrule{3-7}

\multicolumn{1}{c}{\textbf{Method}} & \multicolumn{1}{c}{\bm{$\alpha_{\textbf{train}}$}} & \multicolumn{5}{c}{OOD AUPRC $\uparrow$} \\
\midrule
PA & \cellcolor{blue!20}0.2& 0.864 (.031) & 0.413 (.037) & 0.453 (.028)  & 0.534 (.019) & 0.951 (.009) \\
LinPA & \cellcolor{blue!20}0.2& 0.610 (.082)  & \textbf{0.511 (.011)} & 0.454 (.027)  & 0.536 (.018)  & 0.854 (.049) \\
\rowcolor{green!8}
HyPA & \cellcolor{blue!20}0.2& \textbf{0.919 (.034)} & 0.443 (.070) & \textbf{0.656 (.106)} & \textbf{0.788 (.047)} & \textbf{0.967 (.020)} \\
\midrule[0.01pt]
PA & \cellcolor{red!20}5.0 & 0.581 (.140) &  0.451 (.053)  & 0.497 (.170)  & 0.706 (.021)  & 0.957 (.017) \\
LinPA & \cellcolor{red!20}5.0 & \textbf{0.667 (.087)} &  \textbf{0.518 (.027)}  & \textbf{0.560 (.070)} & 0.566 (.049) & 0.869 (.074) \\
\rowcolor{green!8}
HyPA & \cellcolor{red!20}5.0 & 0.654 (.200) & 0.509 (.043)  & 0.532 (.178) & \textbf{0.751 (.045) } & \textbf{0.978 (.003)} \\
\midrule
\multicolumn{2}{c}{} & \multicolumn{5}{c}{Absolute Slope $\downarrow$} \\
\midrule
PA & \cellcolor{blue!20}0.2 & 0.063 (.013) & 0.225 (.078) & 0.289 (.023) & 0.271 (.012)  & 0.020 (.009)  \\
LinPA & \cellcolor{blue!20}0.2 & 0.080 (.043) & \textbf{0.010 (.013)} & 0.201 (.063) & \textbf{0.029 (.016)} & 0.012 (.006) \\
\rowcolor{green!8}
HyPA & \cellcolor{blue!20}0.2 & \textbf{0.022 (.016)} & 0.169 (.134) & \textbf{0.133 (.075)} & 0.073 (.048) & \textbf{0.011 (.017)} \\
\midrule[0.01pt]
PA & \cellcolor{red!20}5.0 & 0.195 (.066) & 0.124 (.111) & 0.282 (.011) & 0.141 (.014) & 0.014 (.008) \\
LinPA & \cellcolor{red!20}5.0 & \textbf{0.105 (.048)} & \textbf{0.023 (.011)} & \textbf{0.082 (.059)} & \textbf{0.072 (.027)} & 0.011 (.004) \\
\rowcolor{green!8}
HyPA & \cellcolor{red!20}5.0 & 0.174 (.102) & 0.051 (.096)  & 0.211 (.138) & 0.103 (.036) & \textbf{0.002 (.002)} \\
\midrule
\multicolumn{2}{c}{} & \multicolumn{5}{c}{AUSC $\uparrow$} \\
\midrule
PA & \cellcolor{blue!20}0.2 & 0.915 (.015) & 0.565 (.026) & 0.666 (.009) & 0.748 (.010) & 0.967 (.006)  \\
LinPA & \cellcolor{blue!20}0.2 & 0.662 (.053) & 0.526 (.018) & 0.595 (.028) & 0.552 (.026) & 0.863 (.047) \\
\rowcolor{green!8}
HyPA & \cellcolor{blue!20}0.2 & \textbf{0.933 (.014)} & \textbf{0.567 (.023)} & \textbf{0.746 (.036)} & \textbf{0.847 (.009)} & \textbf{0.973 (.011)} \\
\midrule[0.01pt]
PA & \cellcolor{red!20}5.0 & 0.721 (.099) & \textbf{0.550 (.025)} & 0.624 (.037) & 0.815 (.008) &  0.972 (.005)  \\
LinPA & \cellcolor{red!20}5.0 & 0.651 (.036) & 0.526 (.007) & 0.578 (.040) & 0.614 (.068) & 0.868 (.074) \\
\rowcolor{green!8}
HyPA & \cellcolor{red!20}5.0 & \textbf{0.751 (.094)} & 0.546 (.036) & \textbf{0.651 (.036)} & \textbf{0.830 (.017)} & \textbf{0.980 (.001)} \\
\midrule
\multicolumn{2}{c}{} & \multicolumn{5}{c}{Adjusted Integral $\downarrow$} \\
\midrule
PA & \cellcolor{blue!20}0.2 & 0.088 (.016) & 0.442 (.019) & 0.342 (.011) & 0.257 (.010) & 0.034 (.007)  \\
LinPA & \cellcolor{blue!20}0.2 & 0.338 (.053) & 0.493 (.011) & 0.410 (.027) & 0.464 (.025) & 0.214 (.090) \\
\rowcolor{green!8}
HyPA & \cellcolor{blue!20}0.2 & \textbf{0.068 (.014)} & \textbf{0.438 (.019)} & \textbf{0.255 (.037)} & \textbf{0.153 (.009)} & \textbf{0.028 (.012)} \\
\midrule[0.01pt]
PA & \cellcolor{red!20}5.0 & 0.284 (.100) & \textbf{0.452 (.026)} & 0.380 (.037) & 0.191 (.009)  & 0.029 (.005) \\
LinPA & \cellcolor{red!20}5.0 & 0.417 (.068) & 0.501 (.023) & 0.445 (.051) & 0.390 (.070) & 0.166 (.098) \\
\rowcolor{green!8}
HyPA & \cellcolor{red!20}5.0 & \textbf{0.255 (.098)} & 0.455 (.036) & \textbf{0.354 (.038)} & \textbf{0.172 (.018)} & \textbf{0.021 (.001)} \\
\bottomrule[1pt]
\end{tabular}%
}
\label{tab:main_results}

\end{table}
In terms of out-of-distribution predictive performance (OOD AUPRC), HyPA generally attains the highest AUPRC values, particularly when $\alpha_{\text{train}} = 0.2$, with substantial improvements on Hate Speech and Civil Comments. Under the opposite training shift ($\alpha_{\text{train}} = 5.0$), HyPA remains competitive and achieves the best performance on several datasets.

For robustness, measured by Absolute Slope, LinPA often achieves shallower slopes, indicating stronger invariance to varying confounding shift. However, this typically comes at the expense of lower predictive performance due to the linearization (Appendix~\ref{sec:add_analysis}). In contrast, HyPA achieves a more favorable balance between performance and stability, which leads to consistently higher joint metrics (e.g., AUSC and Adjusted Integral). It is worth noting that HyPA consistently outperforms the PA baseline across metrics and configurations.

\section{Analysis of HyPA Results}
We investigate \textbf{how} HyPA achieves strong performance and robustness, providing further empirical analysis of its behavior. We examine how prompt arithmetic steers hidden representations over different $\lambda$s (\S\ref{sec:hidden}) and then propose a mechanistic diagnostic to investigate how HyPA mitigates confounding across datasets (\S\ref{sec:sensitivity}). 
\subsection{Hidden Representation Shift in HyPA}\label{sec:hidden}
\begin{figure}[ht!]
    \centering
    \includegraphics[width=\linewidth]{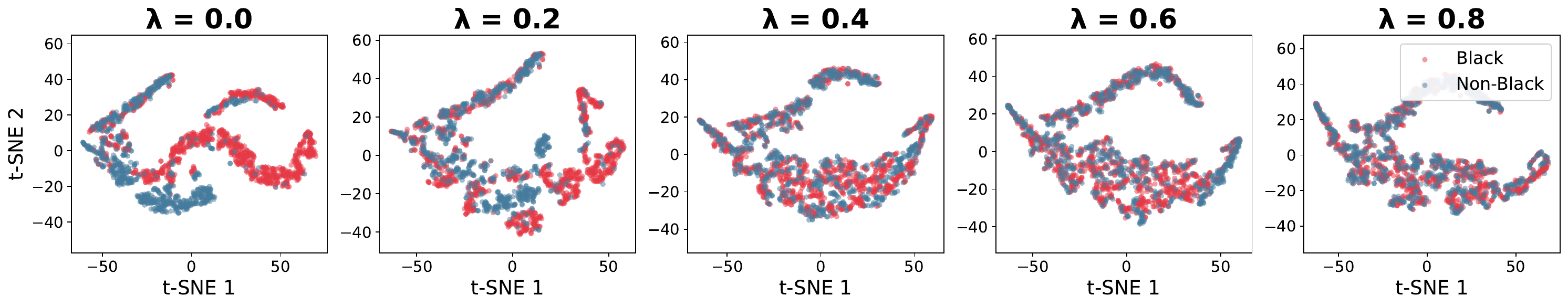}
    \vspace{-5mm}
    \caption{t-SNE plot on the hidden representations for different $\lambda$, colored by the confounder label. The task model is trained on the \textit{Civil Comments} dataset with $\alpha = 0.2$ and evaluated on a randomly sampled test set with $\alpha = 0.7$.}
    \label{fig:tsne-hidden-confounder}
\end{figure}

We begin by visualizing how the hidden representations produced by HyPA evolve as the scaling coefficient $\lambda$ in Equation~\ref{eq:p_debias} varies. Here we provide the results on \textit{Civil Comments} with $\alpha_{\text{train}} = 0.2$ and $\lambda$ ranging between 0.0 and 0.8. As Figure~\ref{fig:tsne-hidden-confounder} suggests, the hidden representations colored by the confounder labels start to blend as $\lambda$ increases, suggesting HyPA diminishes the model's ability to distinguish between confounder groups based on confounder-related features by drawing their representations together in the hidden space.

The t-SNE plot offers diagnostic signals that HyPA is effectively altering the hidden representations of confounder-related features. The gradual blending of the two confounder clusters in this example provides an illustration of HyPA reducing the prominence of confounder-related structure in the hidden representation space. Results of other experimental settings are provided in Appendix~\ref{sec:add_analysis}, where these blending behaviors vary among datasets.
\subsection{How Does HyPA Adjust for Confounders}
\label{sec:sensitivity}

A natural follow-up question is how HyPA improves robustness: does it explicitly remove confounder information, or reshape the learned confounder geometry in the latent space?

Let $h$ denote the hidden representation and define the task direction as 
$W_{\mathrm{task}} = W_U[\mathrm{class1}] - W_U[\mathrm{class2}]$, 
where $W_U$ is the unembedding matrix (i.e. the output embeddings - one for each class). We quantify the model's sensitivity to the confounder as
\begin{equation}
    \mathrm{Sen}(\lambda) = \left| W_{\mathrm{task}} \cdot v_{\text{conf}}(\lambda) \right|,
\end{equation}

where 
$v_{\text{conf}}(\lambda) = \mathbb{E}[h(\lambda)\mid Z{=}1] - \mathbb{E}[h(\lambda)\mid Z{=}0]$
denotes the centroid difference between contextual embedding representations of confounder groups in the hidden space at scaling factor $\lambda$. Here, $W_{\mathrm{task}}$ is fixed, and $v_{\text{conf}}(\lambda)$ serves as a proxy for the confounder direction in the hidden space. This quantity measures the projection of the confounder direction onto the task direction at each $\lambda$.

To assess whether confounder information remains encoded in the representation, we train a linear probe to predict the confounder using the top 50 principal components of $h$ as features. We report probe accuracy averaged over 5-fold cross-validation to ensure robustness.

\begin{figure}[ht!]
    \centering
    \includegraphics[width=\linewidth]{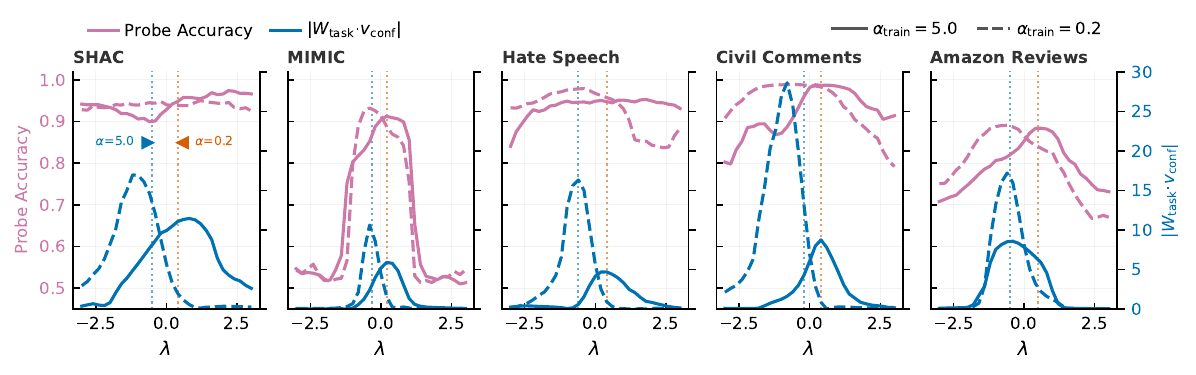}
    \setlength\belowcaptionskip{-3mm}
    \caption{The Sensitivity and Probe Accuracy changes across $\lambda$ sweep. The dotted vertical lines indicate the oracle-optimal $\lambda$ in each setting ($\alpha_{\text{train}}=0.2$ and $\alpha_{\text{train}}=5.0$).}
    
    \label{fig:sens}
\end{figure}

As shown in Figure~\ref{fig:sens}, empirical results across five datasets suggest that, under the same test-time $\alpha$, HyPA mitigates spurious correlations through dataset-dependent mixtures of two effects: reducing the influence of confounder-related variation on task prediction, and weakening confounder information in the representation.

For SHAC and Hate Speech, probe accuracy remains high across the $\lambda$ sweep while sensitivity decreases at extreme $\lambda$ values. This pattern is consistent with a  blocking-like effect: confounder information remains decodable from the hidden representation, while its influence on task prediction is reduced.
In contrast, for MIMIC, sensitivity closely tracks probe accuracy, and both decrease substantially at extreme $\lambda$ values. This behavior is more consistent with representational removal, in which HyPA weakens the confounder information encoded in the hidden space itself.
Civil Comments and Amazon Reviews exhibit intermediate behavior, suggesting that both effects may be present to varying degrees.

Overall, our analysis does not uniquely identify the underlying mechanism, but the joint behavior of sensitivity and probe accuracy provides consistent evidence that HyPA mitigates confounding through dataset-dependent mixtures of blocking-like and removal-like effects.

\section{Comparisons with Other Baselines}\label{sec:other_baseline}

Besides the comparisons within prompt arithmetic variants, we also conduct experiments on the same five benchmarks with other established baseline approaches in domain adaptation or distribution shift, across five random seeds. The included baselines are GroupDRO \citep{sagawa2019distributionally}, JTT \citep{liuJustTrainTwice2021}, DFR \citep{kirichenkoLastLayerReTraining2023}, IRM \citep{arjovsky2019invariant} and DANN \citep{ganinDomainAdversarialTrainingNeural2016}. Besides the existing baselines, we have also investigated two variants of full-parameter task arithmetic: TA and HyTA, corresponding to PA and HyPA, respectively. 

\begin{figure}[ht!]
    \centering
    \includegraphics[width=\linewidth]{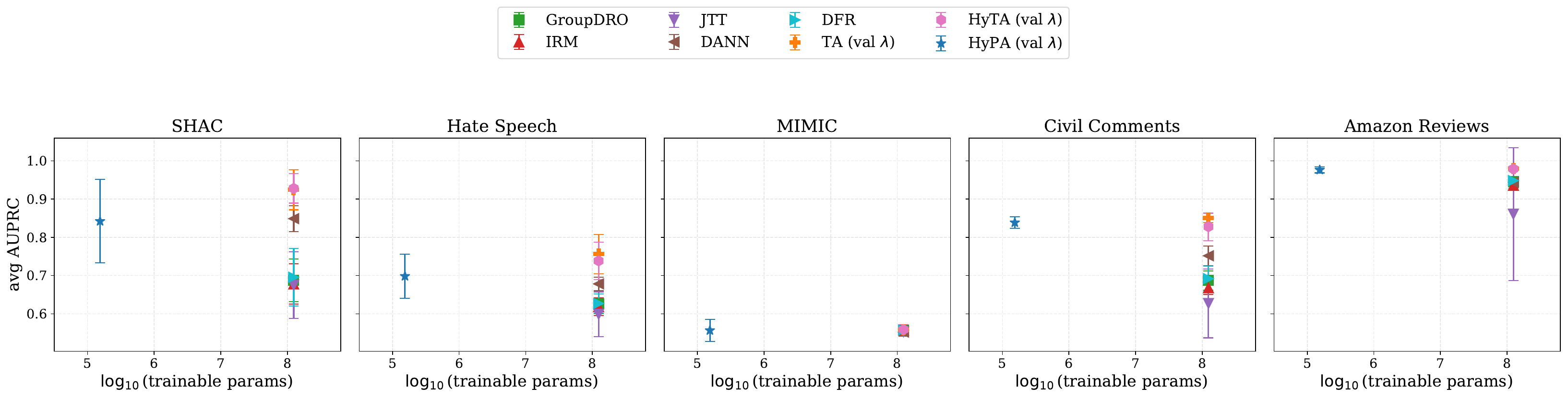}
    \caption{Comparisons of HyPA, Full Param Task Arithmetic, GroupDRO, IRM, JTT, DFR and DANN. Similar to PA and HyPA, we abbreviate Full Param Task Arithmetic approaches as Task Arithmetic (TA) and Hybrid Task Arithmetic (HyTA) respectively.}
    \label{fig:baseline_com}
\end{figure}

Figure~\ref{fig:baseline_com} presents a scatter plot comparing the number of trainable parameters with the average AUPRC across test-time shifts. The results show that HyPA achieves performance comparable to full-parameter task arithmetic methods, including TA and HyTA, while requiring substantially fewer trainable parameters. This suggests that operating in the soft-prompt space can retain much of the robustness benefit of full-model adaptation without incurring its substantially higher parameter and computational costs. HyPA also outperforms established baselines on nearly all benchmarks, demonstrating stronger robustness under confounding shifts. Overall, these results indicate that HyPA offers a favorable performance–efficiency trade-off and provides a practical, parameter-efficient, and scalable alternative to full-parameter debiasing methods.

\section{Ablation Studies}

\begin{wrapfigure}{r}
{0.40\textwidth}
    \centering
    \includegraphics[width=\linewidth]{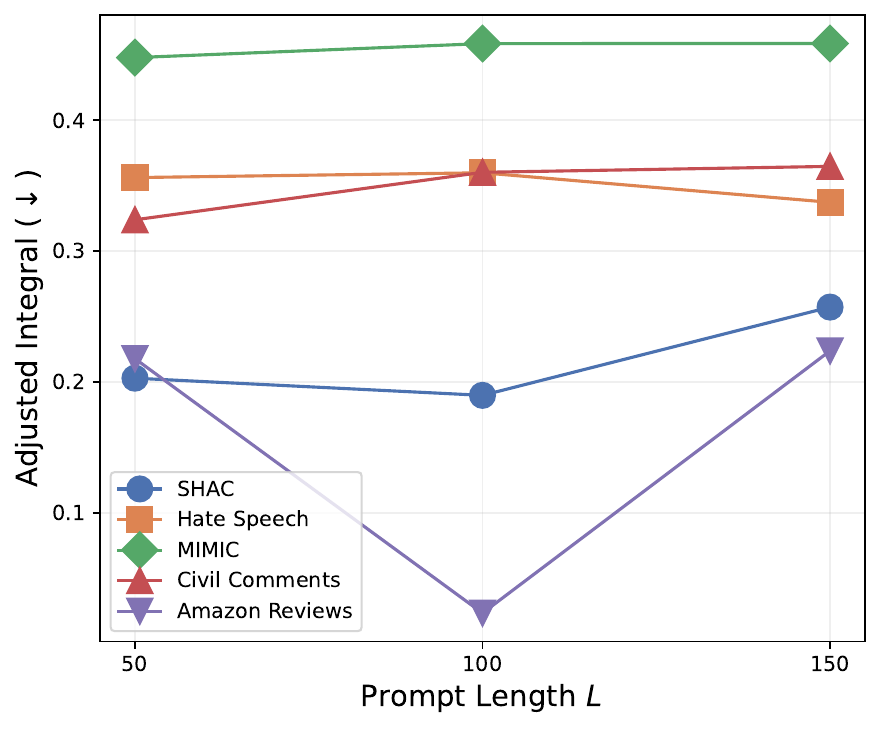}
    {
    \setlength\abovecaptionskip{-3mm}
    \setlength\belowcaptionskip{-3mm}
    \caption{Prompt length ablation with $L = 50, 100, 150$ in five datasets.
    \label{fig:ablat2}
    }
    }
\end{wrapfigure}
\paragraph{Prompt Length.} 
We ablate the number of virtual tokens used in the soft prompts. While the main results are reported with a prompt length of 100, we additionally evaluate prompt lengths of 50 and 150. The results in Figure~\ref{fig:ablat2} show that prompt length does not have a consistent effect on the Adjusted Integral across datasets. In some cases, a longer prompt provides slight improvements, while in others the gains are marginal or even reversed, suggesting that increasing prompt capacity does not uniformly translate to better robustness under confounding shift. The main takeaway from this ablation is that $L=100$ provides a strong and stable default across all five datasets, and overall HyPA is not highly sensitive to prompt length within the ablated range.

\paragraph{Anchor of $\tau_c^{lin}$.}Compared to prior work on task arithmetic with linearization, a key distinction of HyPA is that it anchors the confounder task vector at the fine-tuned weights rather than the pretrained initialization. We further ablate this design choice by recalculating $\tau_c^{lin}$ with respect to the initial weights, and observe a consistent increase in Adjusted Integral (Figure~\ref{fig:ablat1}(a)). This suggests that anchoring at the fine-tuned solution better preserves task performance while improving robustness to confounding shift under most settings. We provide the corresponding ID and OOD AUPRC results in Appendix~\ref{sec:add_analysis} with additional analysis.

{\setlength{\textfloatsep}{-3mm}
\begin{figure}[ht!]
    \centering
    \includegraphics[width=\linewidth]{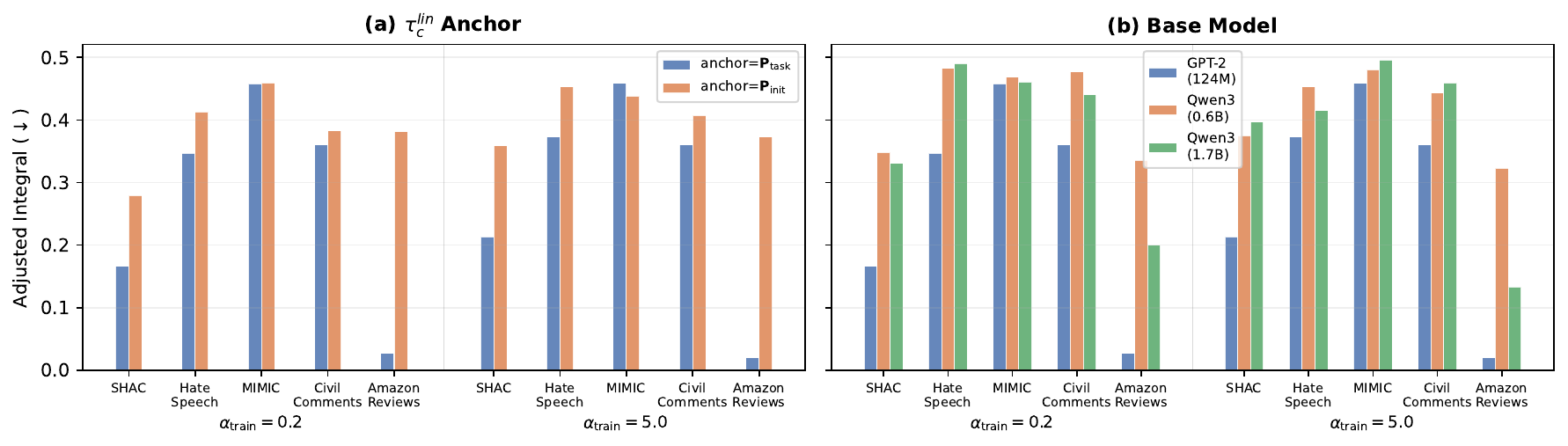}
    \setlength\abovecaptionskip{-3mm}
    \setlength\belowcaptionskip{-3mm}
    \caption{Ablation results for \textbf{(a)}: Anchor of $\tau_c^{lin}$ and \textbf{(b)}: Base Model across five datasets.}
    \label{fig:ablat1}
\end{figure}
}

\paragraph{Base Model.}

We compare GPT-2 (124M) with Qwen3-0.6B, Qwen3-1.7B, and Qwen3-8B \citep{yang2025qwen3technicalreport}. Within the Qwen3 family, absolute debiasing performance generally improves with model size, with Qwen3-8B achieving the strongest results in Table~\ref{tab:hypa_summary}. However, Figure~\ref{fig:ablat1}(b) shows that GPT-2 outperforms Qwen3-0.6B and Qwen3-1.7B across several settings, suggesting that debiasing effectiveness does not scale monotonically across model families.

Table~\ref{tab:hypa_summary} further separates absolute performance from HyPA's marginal benefit over PA. HyPA substantially improves both average AUPRC and Adjusted Integral for Qwen3-0.6B and Qwen3-1.7B, but provides little additional gain at 8B. Thus, larger Qwen models perform better in absolute terms, while the relative advantage of HyPA does not increase consistently with scale. This pattern may reflect architectural differences, model depth, or the use of a fixed prompt length across models, which we leave for future investigation.

\begin{table}[ht!]
\centering
\caption{
Comparisons across the SHAC and Hate Speech dataset under both $\alpha_{\text{train}}$ values.
}
\label{tab:hypa_summary}
\small
\begin{tabular}{lccccccc}
\toprule
& \multicolumn{3}{c}{Average AUPRC $\uparrow$}
& \multicolumn{3}{c}{Adjusted Integral $\downarrow$} \\
\cmidrule(lr){2-4}
\cmidrule(lr){5-7}
Qwen3 Model
& PA
& HyPA
& $\Delta$ value
& PA
& HyPA
& $\Delta$ ratio \\
\midrule
0.6B
& 0.8178
& \textbf{0.8433}
& \textbf{+0.0255}
& 0.1855
& \textbf{0.1580}
& \textbf{-14.8\%} \\

1.7B
& 0.8070
& \textbf{0.8450}
& \textbf{+0.0380}
& 0.1980
& \textbf{0.1583}
& \textbf{-20.1\%} \\

8B
& \textbf{0.9128}
& 0.9123
& $-0.0005$
& 0.0903
& \textbf{0.0895}
& -0.8\% \\
\bottomrule
\end{tabular}
\end{table}

\section{Related Work}

\paragraph{Distribution Shift.} 

Machine learning models and data-driven systems often suffer performance degradation under distribution shift at deployment, where the joint distribution of inputs and labels differs from that of the training environment. Prior work has investigated various forms of shift, including domain generalization \citep{zhou2022domain}, covariate shift \citep{shimodaira2000improving, ruan2021optimal}, label shift \citep{lipton2018detecting}, and subpopulation shift \citep{koh2021wilds, yang2023change}.
In this work, we focus on \textit{confounding shift} \citep{landeiro2018robust, ding2024backdoor}, a structured form of subpopulation shift in which a confounder $Z$ causally influences both the input $X$ and the label $Y$. We place particular emphasis on developing parameter-efficient methods to mitigate confounding shift in large neural language models, leveraging prompt linearization to enable effective task arithmetic.

\paragraph{Bias Mitigation.} Our work addresses bias mitigation under confounding shifts. Existing approaches are typically categorized into pre-processing, in-processing, and post-processing methods \citep{angwin2016machine}. Pre-processing methods modify the training data through feature editing \citep{feldman2015certifying, calmon2017optimized} or reweighting \citep{kamiran2012data}, while in-processing approaches incorporate fairness constraints directly into model optimization \citep{agarwal2018reductions, zhang2018mitigating}. Post-processing methods instead adjust the outputs of a trained model to satisfy fairness criteria \citep{hardt2016equality, pleiss2017fairness}, without requiring access to or retraining of the underlying model.
HyPA falls into the post-processing category and operates without modifying the original training pipeline. Compared to prior post-processing methods, which typically rely on output calibration, threshold adjustment or weight merging \citep, HyPA leverages parameter-efficient prompt composition to intervene at the representation level. This enables more flexible mitigation of confounding effects while preserving the efficiency and modularity of post-hoc approaches, making it particularly suitable for large language models where full retraining is impractical.

\section{Conclusion}

We presented Hybrid Prompt Arithmetic (HyPA), a parameter-efficient method for improving robustness under confounding shift. By combining non-linear task prompt tuning with linearized confounder tuning, HyPA balances expressivity and disentanglement and allows prompt arithmetic to mitigate spurious correlations while preserving task performance.

Across multiple datasets and shift settings, HyPA consistently achieves a stronger robustness-performance trade-off than prompt-arithmetic baselines based solely on non-linear or linearized fine-tuning. Our analysis further suggests that HyPA operates in a dataset-dependent manner: in some cases, HyPA primarily reduces the influence of confounder-related variation on task prediction, whereas in others it is also associated with weaker confounder information in the representation itself. These results position HyPA as a simple and effective post-hoc intervention for improving robustness in the evaluated prompt-tuned language model setting.

Despite these promising results, several limitations remain. First, our analysis is primarily empirical, and a stronger theoretical account is needed to characterize when and why
hybrid prompt arithmetic can isolate useful confounder directions. Second, although
HyPA acts only in soft prompt space, the precise way prompt updates propagate through
the frozen backbone and affect the unembedding layer remains insufficiently understood.
Third, as discussed in Appendix~\ref{sec:select}, model selection under unknown test-time shift remains challenging, and developing validation criteria that better align with downstream robustness is an important direction for future work.

\section*{Ethics Statement}
This work addresses robustness under confounding shift in settings where spurious correlations can lead to brittle or unfair predictions. A potential benefit of HyPA is reduced reliance on confounder-related signals, which may improve robustness and reduce group disparities under distribution shift. However, such improvements do not guarantee fairness or robustness in deployment. The effectiveness of any debiasing method depends on the quality of confounder annotations, the data distribution, and the target environment. In particular, datasets drawn from clinical and online text domains may contain historical biases, annotation artifacts, and subgroup imbalance, and errors in these settings can cause disproportionate harm. Our results should therefore be interpreted as evidence of improved robustness on the studied benchmarks, not as a complete solution to bias or fairness concerns.



\section*{Acknowledgments}
This work was supported by U.S. National Library of Medicine grant (R01LM014056).

\bibliography{colm2026_conference}
\bibliographystyle{colm2026_conference}
\newpage
\appendix

{\Large{Appendix}}
\section{HyPA algorithm}\label{alg:hypa}

\begin{algorithm}[ht!]
\caption{Hybrid Prompt Arithmetic (HyPA)}
\label{hypa_alg}
\begin{algorithmic}[1]
\Require Frozen language model $f_{\theta}$, initial soft prompt $\mathbf{P}_0$, dataset $\mathcal{D}=\{(x_i,y_i, z_i)\}_{i=1}^N$, scale $\lambda$
\Ensure De-confounded soft prompt $\mathbf{P}_{\text{debias}}$
\Statex
\textbf{Phase 1: Non-linear task tuning}
\State Initialize task prompt $\mathbf{P} \gets \mathbf{P}_0$
\State Optimize $\mathbf{P}$ on $\mathcal{D}$ with the full non-linear model:
\[
\mathbf{P}_t \gets \arg\min_{\mathbf{P}} \; \mathbb{E}_{(x,y)\sim\mathcal{D}}
\left[\mathcal{L}\bigl(f_{\theta}(x;\mathbf{P}), y\bigr)\right]
\]
\State Compute the task vector:
\[
\tau_t \gets \mathbf{P}_t - \mathbf{P}_0
\]

\Statex
\textbf{Phase 2: Linearized confounder tuning around $\mathbf{P}_t$}
\State Initialize confounder update $\tau_c^{\mathrm{lin}} \gets \mathbf{0}$
\State Linearize the model at $\mathbf{P}_t$:
\[
f_{\theta}^{\mathrm{lin}}(x;\mathbf{P}_t+\tau)
\;=\;
f_{\theta}(x;\mathbf{P}_t)
+
\nabla_{\mathbf{P}} f_{\theta}(x;\mathbf{P}_t)^{\top}\tau
\]
\State Optimize the linearized confounder objective:
\[
\tau_c^{\mathrm{lin}}
\gets
\arg\min_{\tau}\;
\mathbb{E}_{(x,z)\sim\mathcal{D}}
\left[\mathcal{L}_c\bigl(f_{\theta}^{\mathrm{lin}}(x;\mathbf{P}_t+\tau), z\bigr)\right]
\]

\Statex
\textbf{Prompt Arithmetic}
\State Compose the de-confounded prompt:
\[
\mathbf{P}_{\text{debias}}
\gets
\mathbf{P}_0 + \tau_t + \lambda \,\tau_c^{\mathrm{lin}}
\]

\State \Return $\mathbf{P}_{\text{debias}}$
\end{algorithmic}
\end{algorithm}
\section{Additional Analysis Results}\label{sec:add_analysis}
\subsection{Bias from Oracle Selection}\label{sec:select}
In general, the test-time distribution is assumed to be unavailable in distribution shift settings during training \citep{gulrajani2020search, yang2023change}, preventing selection of optimal hyperparameters for test-time performance. To assess the potential impact of this limitation, we demonstrate an oracle setting where $\lambda$ is chosen by minimizing the Adjusted Integral on the test distribution. Table~\ref{tab:oracle_comparison} reports the results on SHAC with $\alpha_{\text{train}}=5.0$. HyPA remains the best-performing method even under oracle selection, indicating that its advantage is not solely due to hyperparameter tuning but reflects improved robustness to confounding shift. This gap also raises questions about how to find a better proxy for model selection when the test environment is inaccessible.
\begin{table}[ht]
\centering
\caption{Oracle hyperparameter selection comparison on SHAC ($\alpha_{\text{train}}=5.0$, averaged over 5 seeds). Grey values correspond to $\lambda$ selected using the validation set.}
\resizebox{\textwidth}{!}{%
\begin{tabular}{lccccc}
\toprule
Method & Adjusted Integral $\downarrow$ & AUSC $\uparrow$ & OOD AUPRC $\uparrow$ & ID AUPRC $\uparrow$ \\
\midrule
PA &  0.24 (.08)/\textcolor{gray}{0.28 (.10)}& 0.76 (.08)/\textcolor{gray}{0.72 (.10)} & 0.58 (.14)/\textcolor{gray}{0.58 (.14)} & \textbf{0.87 (.08)}/\textcolor{gray}{0.85 (.07)} \\
LinPA & 0.39 (.05)/\textcolor{gray}{0.42 (.07)} & 0.65 (.03)/\textcolor{gray}{0.65 (.04)} & 0.53 (.06)/\textcolor{gray}{0.67 (.09)} & 0.63 (.05)/\textcolor{gray}{0.62 (.09)} \\
HyPA &  \textbf{0.22 (.07)}/\textcolor{gray}{0.26 (.10)} & \textbf{0.78 (.07)}/\textcolor{gray}{0.75 (.09)} & \textbf{0.66 (.11)}/\textcolor{gray}{0.65 (.20)} & 0.84 (.06)/\textcolor{gray}{0.84 (.10)} \\
\bottomrule
\end{tabular}
}
\label{tab:oracle_comparison}
\end{table}

\subsection{Examining Group Fairness Metrics}\label{sec:other_met}
\begin{figure}[ht!]
    \centering
    \includegraphics[width=\linewidth]{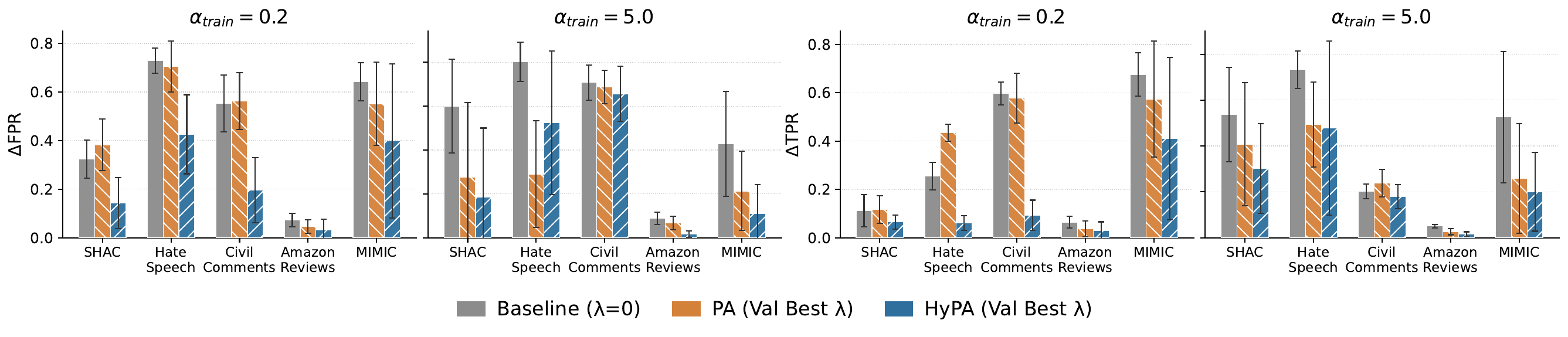}
    \caption{Comparisons of group fairness metrics in Out of Distribution setting.}
    \label{fig:fair_met}
\end{figure}

In addition to the mechanism analysis from hidden representations, we report point evaluations in the OOD setting using two widely adopted fairness metrics: $\Delta \text{FPR}$ and $\Delta \text{TPR}$. They are calculated as the absolute difference in False Positive Rate and True Positive Rate between two sensitive groups (in this case, the groups are determined by the confounding variable), accordingly. As shown in Figure~\ref{fig:fair_met}, across five datasets, HyPA consistently achieves lower $\Delta \text{FPR}$ and $\Delta \text{TPR}$ compared to both baselines. The results indicate that HyPA provides better control over error disparities between confounder groups in OOD settings.
We do not report Statistical Parity (SP) in this context, as $P(Y \mid Z)$ is inherently imbalanced in the OOD test sets, making SP comparisons less meaningful and potentially misleading.

\subsection{Additional Results on Anchor Ablations}
\begin{figure}[ht!]
    \centering
    \includegraphics[width=\linewidth]{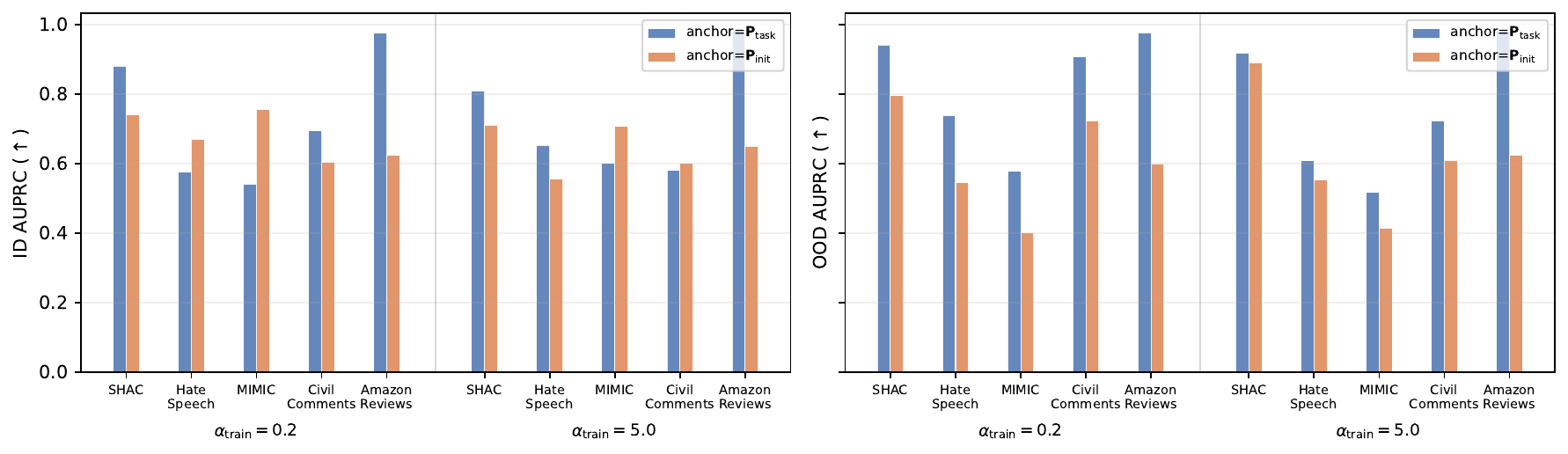}
    \caption{In-Distribution AUPRC and Out-of-Distribution AUPRC across five datasets.}
    \label{fig:add_anchor}
\end{figure}

In Figure~\ref{fig:add_anchor}, we observe that anchoring HyPA at $\mathbf{P}{\text{task}}$ consistently yields better OOD AUPRC across all configurations. However, anchoring at the initialization weights leads to improvements in ID AUPRC under certain settings. Although the results are reported using the oracle $\lambda$ selected by Adjusted Integral, anchoring at initialization consistently outperforms the task prompt anchor on the MIMIC dataset when considered alongside Figure~\ref{fig:ablat1}(a), particularly at $\alpha{\text{train}} = 5.0$.

Revisiting the observations in Figure~\ref{fig:sens}, we hypothesize that in datasets where task-relevant and confounding features are strongly entangled, anchoring at initialization produces a confounder task vector that is less influenced by task-specific signals. Consequently, adding or subtracting this vector has a reduced impact on task performance, leading to weaker deconfounding effects in the resulting task arithmetic composition. We note that this hypothesis remains preliminary and requires further empirical validation.


\subsection{Non-linear Advantage}
\begin{figure}[ht]
    \centering
    \includegraphics[width=0.5\linewidth]{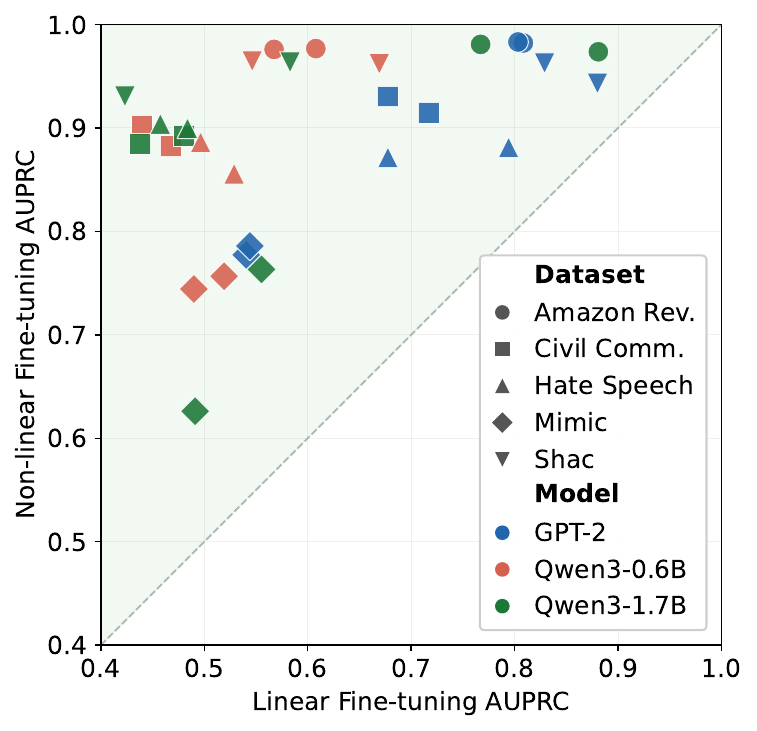}
    \caption{Evidence of the advantage of non-linear fine-tuning across datasets and models.}
    \label{fig:non-linear}
\end{figure}

Figure~\ref{fig:non-linear} visualizes the advantage of non-linear fine-tuning over linear fine-tuning for task prediction across different models and datasets. Non-linear fine-tuning consistently obtains higher task-prediction AUPRC.

\subsection{Additional t-SNE visualizations}

\begin{figure}[ht!]
    \centering
    \includegraphics[width=\linewidth]{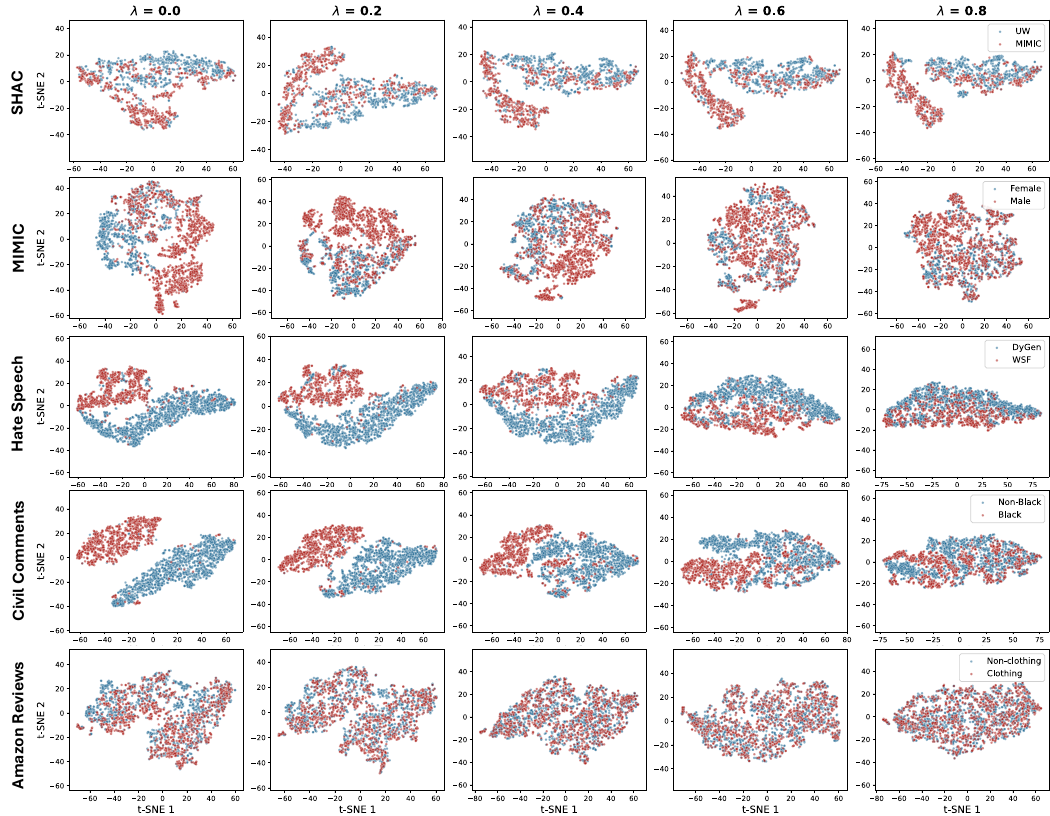}
    \caption{t-SNE plot for $\alpha_{\text{train}}= 0.2$ configuration across datasets. Points are labeled by their corresponding confounder groups.}
    
    \label{fig:tsne-0.2}
\end{figure}

\begin{figure}[ht!]
    \centering
    \includegraphics[width=\linewidth]{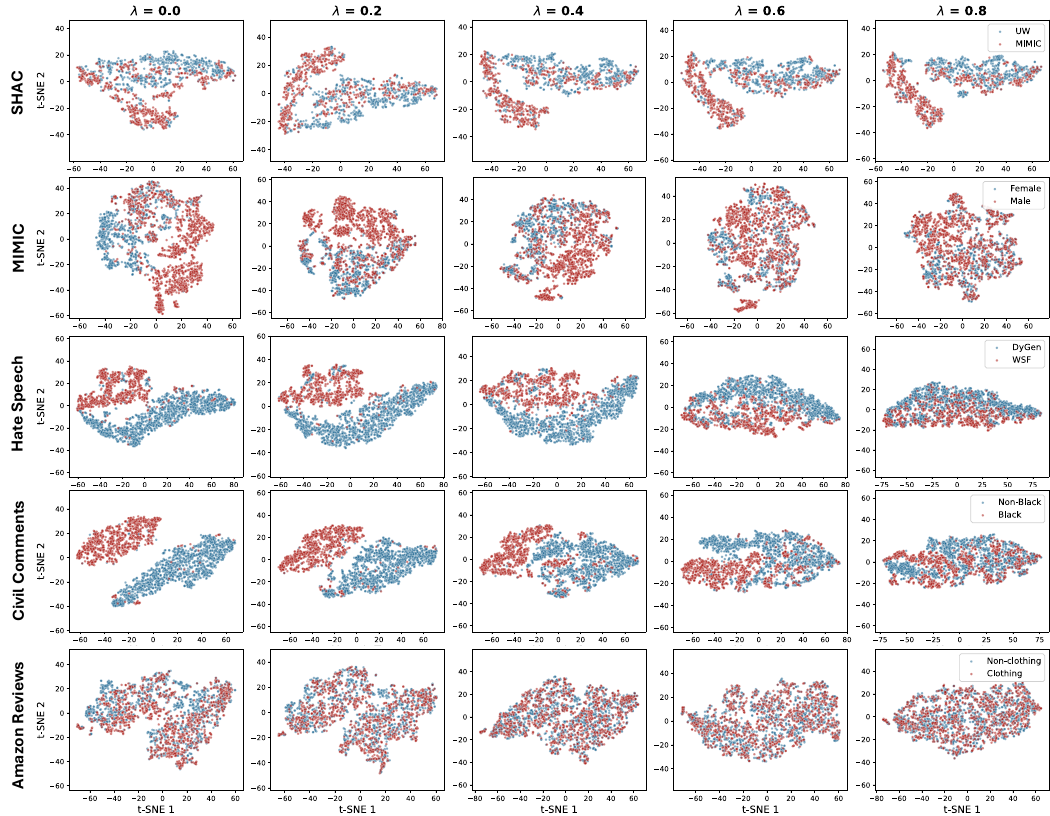}    
    \caption{t-SNE plot for $\alpha_{\text{train}}= 5.0$ configuration across datasets. Points are labeled by their corresponding confounder groups.}
    \label{fig:tsne-5.0}
\end{figure}

In Figure~\ref{fig:tsne-0.2} and Figure~\ref{fig:tsne-5.0}, we display the full panel of t-SNE visualizations of the hidden space given the confounder label in each dataset. The scatter plot is generated via 2000 random samples from the dataset. PCA dimension reduction is applied first to extract the first 50 principal components for efficient computations. From the results, we observe that tuning $\lambda$ more effectively blends the two confounder groups in datasets where the original clusters are well separated (e.g., MIMIC and Hate Speech). In contrast, its impact is less pronounced in datasets where the two sources are already intermixed in the t-SNE space, leaving limited room for further alignment.

\section{Dataset Details}\label{sec:dataset}
We evaluate on five datasets in our experiments, each adapted from prior work.

\paragraph{SHAC \citep{lybarger2021annotating}}
The Social History Annotation Corpus (SHAC) consists of clinical notes collected from two institutions: the University of Washington Medical Center and MIMIC-III. The primary task is to identify substance use information from clinical text. In our experiments, we define the prediction label as the presence of drug abuse and treat the data source (i.e., institution) as the confounder.

\paragraph{Hate Speech \citep{vidgen2021learning, de2018hate}}
We utilize a hate speech detection dataset curated by \citet{dingTailoringTaskArithmetic2025}, which aggregates samples from two distinct sources: (1) synthetically generated text and (2) posts from a white supremacist forum. The task is to detect hate speech, while the data source serves as the confounder.

\paragraph{Civil Comments \citep{borkan2019nuanced}}
Civil Comments is a large collection of user comments from online articles. We follow the preprocessing procedure in \textsc{WILDS} \citep{koh2021wilds}. The task is toxicity prediction from text. We use demographic identity mentions as proxies for potential confounding factors, as such attributes are often spuriously correlated with toxicity labels. In particular, we select mentions of \textit{Black} identity as the confounder.

\vspace{-2mm}
\paragraph{MIMIC \citep{johnson2016mimic}}
MIMIC-Note consists of clinical notes recorded within the first 48 hours of a hospital stay, sourced from the MIMIC-III database. We use curated notes to predict discharge or mortality outcomes. Patient sex (male vs.\ female) is treated as the confounder.
\vspace{-2mm}
\paragraph{Amazon Reviews \citep{ni2019justifying}}
This dataset includes product reviews with ratings, textual content, and metadata. We focus on the review text and define the prediction task as sentiment classification (positive vs.\ negative). The product category (apparel vs.\ non-apparel) is used as the confounder.

Below we present the statistics of each dataset. To construct the different $\alpha$ configurations, we sample from the original dataset to fulfill the distribution requirements. For small datasets like SHAC, MIMIC and Hate Speech, we use a ratio of 0.7/0.1/0.2 to form the training, evaluation, and test set. For larger datasets like Amazon Reviews and Civil Comments, we use a fixed number of 10,000/5,000/5,000 to split.
\begin{table}[ht!]
\centering
\caption{Dataset Statistics}
\label{table:dataset_stats}
\begin{tabular}{lrrrr}
\toprule
\textbf{Dataset} & \textbf{$N$} & \textbf{P(Y=1)} & \textbf{P(Z=1)}  \\
\midrule
SHAC & 4,405 & 0.3203 & 0.4261  \\
MIMIC & 25,880 & 0.1381 & 0.5720 \\
Hate Speech & 51,847 & 0.4508 & 0.2064 \\
Amazon Reviews & 29,557,446 & 0.8610 & 0.3939 \\
Civil Comments & 447,998 & 0.0804 & 0.0341 \\
\bottomrule
\end{tabular}
\end{table}

\section{Training Details}\label{sec:training}
\subsection{LM Head for Binary Classification}\label{sec:frozen_lm}
In our setup, we restrict the frozen backbone to decoder-only transformer architectures and repurpose the language model (LM) head, also referred to as the unembedding matrix, for binary classification. Specifically, let \textit{class1} and \textit{class2} denote two label tokens associated with the target classes for a given input text sequence $X$. During training and inference, we retain only the logits corresponding to these label tokens and use them to compute the classification loss.

\begin{equation}
\mathbb{P}(\text{class1} \mid X)
= \frac{\exp(\mathbf{s}^{\text{class1}})}
{\exp(\mathbf{s}^{\text{class1}}) + \exp(\mathbf{s}^{\text{class2}})},
\end{equation}

where $\mathbf{s}^{\text{class1}}$ and $\mathbf{s}^{\text{class2}}$
 denote the output logits produced by the LM head for the corresponding class tokens. The predicted label is the class token with the higher probability. Throughout soft prompt training, all preexisting model parameters, including the language model head, are frozen. Consequently, task adaptation occurs solely through the soft prompt embeddings, eliminating the need for an additional classification head.  We adopt this setup because the pretrained LM head provides a semantically structured decoding space already aligned with next-token prediction and instruction-following behavior. By mapping class labels to meaningful tokens, the model can exploit this pretrained decision structure directly, rather than introducing and optimizing a separate classifier. Because the entire backbone remains frozen, adaptation is driven exclusively by soft prompt embeddings that modulate internal representations. This formulation preserves the model’s original prediction structure and supports task-specific conditioning, providing a stable and parameter-efficient framework for classification. Of note, it also permits modification of model capabilities through task arithmetic operations on the prompt embeddings alone.

\subsection{Training Hyperparameters}

\begin{table}[ht!]
\centering
\caption{Training hyperparameters for main results.}
\label{tab:training_hyperparams}
\small
\begin{tabular}{ll}
\toprule
\textbf{Setting} & \textbf{Value} \\
\midrule
Base model & GPT-2 \\
Prompt method & Prompt tuning \\
Prompt length & 100 \\
Max sequence length & 256 \\
Optimizer & AdamW \\
Scheduler & Linear decay with 10-step warmup \\
Learning rate (task / confounder) & $10^{-3}$ / $10^{-3}$ \\
Precision & BF16 \\
Batch size (train / eval) & 8 / 16 per device \\
Epochs (task/confounder)  & 100 / 100 \\
Logging / evaluation frequency & 20 / 40 steps \\
Random seed & 0,1,2,3,4 \\
$\alpha_{\text{train}}$ values & $0.2, 5.0$ \\
Data split (train / eval / test) & 70\% / 10\% / 20\% or 10k / 5k / 5k\\
\bottomrule
\end{tabular}
\end{table}




\end{document}